# Deep Mixtures of Factor Analysers

**Yichuan Tang**                                TANG@CS.TORONTO.EDU
**Ruslan Salakhutdinov**                        RSALAKHU@CS.TORONTO.EDU
**Geoffrey Hinton**                             HINTON@CS.TORONTO.EDU
Department of Computer Science, University of Toronto, Toronto, Ontario, CANADA

## Abstract

An efficient way to learn deep density models that have many layers of latent variables is to learn one layer at a time using a model that has only one layer of latent variables. After learning each layer, samples from the posterior distributions for that layer are used as training data for learning the next layer. This approach is commonly used with Restricted Boltzmann Machines, which are *undirected* graphical models with a single hidden layer, but it can also be used with Mixtures of Factor Analysers (MFAs) which are *directed* graphical models. In this paper, we present a greedy layer-wise learning algorithm for Deep Mixtures of Factor Analysers (DMFAs). Even though a DMFA can be converted to an equivalent shallow MFA by multiplying together the factor loading matrices at different levels, learning and inference are much more efficient in a DMFA and the sharing of each lower-level factor loading matrix by many different higher level MFAs prevents overfitting. We demonstrate empirically that DM-FAs learn better density models than both MFAs and two types of Restricted Boltzmann Machine on a wide variety of datasets.

## 1. Introduction

Unsupervised learning is important for revealing structure in the data and for discovering features that can be used for subsequent discriminative learning. It is also useful for creating a good prior that can be used for tasks such as image denoising and inpainting or tracking animate motion.

A recent latent variable density model based on Markov Random Fields is the Gaussian Restricted Boltzmann Machine (GRBM) (Hinton & Salakhutdinov, 2006). A GRBM can be viewed as a mixture of diagonal Gaussians with the number of components exponential in the number of hidden variables, but with a lot of parameter sharing between the exponentially many Gaussians. In (Hinton et al., 2006), it was shown that a trained RBM model can be improved by using a second RBM to create a model of the "aggregated posterior" (Eq. 9) of the first RBM, where the aggregated posterior is the equally weighted average of the posterior distributions over the hidden units of the first RBM for each training case. The second RBM is then used to replace the prior over the hidden units of the first RBM that is implicitly defined by the weights and biases of the first RBM. With mild assumptions on how training is performed at the higher layer, it was proven that a variational lower bound on the log-likelihood is guaranteed to improve. The second level RBM can do a better job of modeling the first RBM's aggregated posterior than the first level RBM because its parameters are not also being used to model the conditional distribution of the data given the states of the units in the first hidden layer.

A rival model for real-valued high-dimensional data is the Mixture of Factor Analyzers (MFA) (Ghahramani & Hinton, 1996). MFAs simultaneously perform clustering and dimensionality reduction of the data by making locally linear assumptions (Verbeek, 2006). Unlike RBMs, MFAs are directed graphical models where a multivariate standard normal prior is specified for the latent factors for all components. Learning typically uses the EM algorithm to maximize the data log-likelihood. Each FA in the mixture has an isotropic Gaussian prior over its factors and a Gaussian posterior for each training case, but when the posterior is aggregated over many training cases it will typically be non-Gaussian. We can, therefore, improve a variational lower bound on the log probability of the training data by replacing the prior of each FA by a separate, second-level MFA that learns to model the aggregated posterior of that FA better than it is modeled by an isotropic Gaussian. Empirically, the average test log-likelihood also increases for models of both low and high-dimensional data.

While it is true that a two layer MFA can be collapsed back into a standard one layer MFA, learning the two models is nevertheless quite different due to the sharing of factor loadings among the second layer components of the Deep MFA. Parameter sharing helps to reduce overfitting and greatly reduces the computational cost of learning. The EM algorithm also benefits from an easier objective function due to the greedy layer-wise learning, so it is less likely to get stuck in poor local optima.

Multilayer factor analysis was also part of the model in (Chen et al., 2011). However, that work mainly focused on learning convolutional features with nonparametric Bayesian priors on the *parameters*. By using max-pooling and decimation of the first layer factors, their model was designed to learn discriminative features, rather than a top-down generative model of pixel values.

## 2. Mixture of Factor Analysers

Factor analysis was first introduced in psychology as a latent variable model to find the "underlying factor" behind covariates. The latent variables are called factors and are of lower dimension than the covariates. Factor analyzers are linear models as the factor loadings span a linear subspace within the vector space of the covariates. To deal with non-linear data distributions, Mixtures of Factor Analyzers (MFA) (Ghahramani & Hinton, 1996) can be used. MFAs approximate nonlinear manifolds by making local linear assumptions.

Let $\mathbf{x} \in \mathbb{R}^D$ denote the $D$-dimensional data, $\{\mathbf{z} \in \mathbb{R}^d : d \leq D\}$ denote the $d$-dimensional latent variable, and $c \in \{1, \ldots, C\}$ denote the component indicator variable of $C$ total components. The MFA is a directed generative model, defined as follows:

$$p(c) = \pi_c, \quad \sum_{c=1}^{C} \pi_c = 1, \tag{1}$$

$$p(\mathbf{z}|c) = p(\mathbf{z}) = \mathcal{N}(\mathbf{z}; \mathbf{0}, \mathbf{I}), \tag{2}$$

$$p(\mathbf{x}|\mathbf{z}, c) = \mathcal{N}(\mathbf{x}; \mathbf{W}_c \mathbf{z} + \boldsymbol{\mu}_c, \boldsymbol{\Psi}_c), \tag{3}$$

where $\mathbf{I}$ is the $d \times d$ identity matrix. The parameters of the $c$-th component include a mixing proportion $\pi_c$, a factor loading matrix $\mathbf{W}_c \in \mathbb{R}^{D \times d}$, mean $\boldsymbol{\mu}_c$, and a diagonal matrix $\boldsymbol{\Psi}_c \in \mathbb{R}^{D \times D}$, which represents the independent noise variances for each of the variables.

By integrating out the latent variable $\mathbf{z}$, a MFA model becomes a mixture of Gaussians with constrained covariance:

$$p(\mathbf{x}|c) = \int_{\mathbf{z}} p(\mathbf{x}|\mathbf{z}, c) p(\mathbf{z}|c) d\mathbf{z} = \mathcal{N}(\mathbf{x}; \boldsymbol{\mu}_c, \Gamma_c) \tag{4}$$

$$\Gamma_c = \mathbf{W}_c \mathbf{W}_c^\mathsf{T} + \boldsymbol{\Psi}_c$$

$$p(\mathbf{x}) = \sum_{c=1}^{C} \pi_c \, \mathcal{N}(\mathbf{x}; \boldsymbol{\mu}_c, \Gamma_c). \tag{5}$$

### Inference

For inference, we are interested in the posterior:

$$p(\mathbf{z}, c|\mathbf{x}) = p(\mathbf{z}|\mathbf{x}, c) p(c|\mathbf{x}) \tag{6}$$

The posterior over the components can be found using Bayes rule:

$$p(c|\mathbf{x}) = \frac{p(\mathbf{x}|c) p(c)}{\sum_{\gamma=1}^{C} p(\mathbf{x}|\gamma) p(\gamma)} \tag{7}$$

Given component $c$, the posterior over the latent factors is also a multivariate Gaussian:

$$p(\mathbf{z}|\mathbf{x}, c) = \mathcal{N}(\mathbf{z}; \mathbf{m}_c, \mathbf{V}_c^{-1}), \tag{8}$$

where

$$\mathbf{V}_c = \mathbf{I} + \mathbf{W}_c^\mathsf{T} \boldsymbol{\Psi}_c^{-1} \mathbf{W}_c,$$
$$\mathbf{m}_c = \mathbf{V}_c^{-1} \mathbf{W}_c^\mathsf{T} \boldsymbol{\Psi}_c^{-1} (\mathbf{x} - \boldsymbol{\mu}).$$

Maximum likelihood learning of a MFA model is straightforward using the EM algorithm. During the E-step, Eqs. 7, 8 are used to compute the posterior over the latent variables given the current setting of the model parameters. During the M-step, the expected complete-data log-likelihood is maximized with respect to the model parameters $\theta = \{\pi_c, \mathbf{W}_c, \boldsymbol{\mu}_c, \boldsymbol{\Psi}_c\}_{c=1}^{C}$:

$$\mathbb{E}_{p(\mathbf{z}, c|\mathbf{x}; \theta_{old})} [\log p(\mathbf{x}, \mathbf{z}, c; \theta)]$$

## 3. Deep Mixtures of Factor Analysers

After MFA training reaches convergence, the model can be improved by increasing the number $C$ of mixture components or the dimensionality $d$ of the latent factors per component. This amounts to adjusting the conditional distributions $p(\mathbf{x}|\mathbf{z}, c)$. However, as we demonstrate in our experimental results, this approach

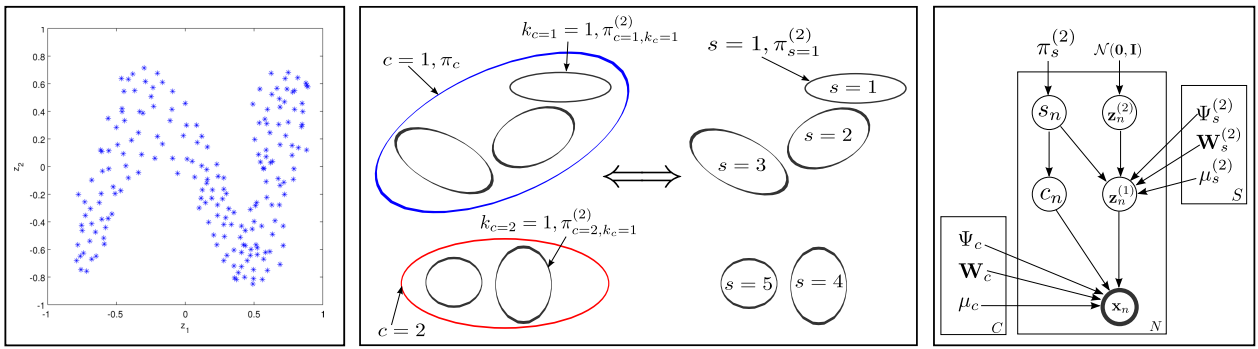

Figure 1. **Left**: The aggregated posterior of a *single* component may not be Gaussian distributed. **Middle**: Illustration of our model for 2D data with each ellipse representing a Gaussian component. The first layer MFA has two components colored blue ($c = 1$) and red ($c = 2$). Their mixing proportions are given by $\pi_c$. For the blue component, we further learn a second layer MFA with three components. For the red component, we learn a separate second layer MFA with two components. We also introduce the second layer component indicator variable $k_c = 1, \dots, K_c$, where $K_c$ is the total number of the second layer components associated with the first layer component $c$. $K_c$ is specific to the first layer component and need not be same for all $c$. In our example, $K_1 = 3$ and $K_2 = 2$. **Right**: Graphical model of a DMFA. Best viewed in color.

quickly leads to overfitting, particularly when modeling high-dimensional data.

An alternative is to replace the standard multivariate normal prior on the latent factors: $p(\mathbf{z}|c) = \mathcal{N}(\mathbf{0}, \mathbf{I})$. The "aggregated posterior" is the empirical average over the data of the posteriors over the factors: $\frac{1}{N} \sum_{n=1}^{N} \sum_{c=1}^{C} p(\mathbf{z}_n, c|\mathbf{x}_n)$ and a component-specific aggregated posterior is:

$$\frac{1}{N} \sum_{n=1}^{N} p(\mathbf{z}_n, c_n = c|\mathbf{x}_n) \tag{9}$$

If each factor analyser in the mixture was a perfect model of the data assigned to it, the component-specific aggregated posterior would be distributed according to an isotropic Gaussian, but in practice, it is non-Gaussian. Figure 1 (left panel) shows a component-specific aggregated posterior (with $d = 2$), which is highly non-Gaussian. In this case, we wish to replace a simple standard normal prior by a more powerful MFA prior:

$$p(\mathbf{z}|c) = \text{MFA}(\boldsymbol{\theta}_c^{(2)}) \tag{10}$$

Here, $\boldsymbol{\theta}_c^{(2)}$ emphasizes that the new MFA's parameters are at the second layer and are specific to component $c$ of the first layer MFA.

More concretely, the variational lower bound on the log-likelihood of the model given data $\mathbf{x}$ is:

$$\mathcal{L}(\mathbf{x}; \boldsymbol{\theta}) = \sum_{c=1}^{C} \int_{\mathbf{z}} q(\mathbf{z}, c|\mathbf{x}; \boldsymbol{\theta}) \log p(\mathbf{x}, \mathbf{z}, c; \boldsymbol{\theta}) d\mathbf{z} + \mathcal{H}(q)$$

$$= \sum_{c=1}^{C} \int_{\mathbf{z}} q(\mathbf{z}, c|\mathbf{x}; \boldsymbol{\theta}) \Big\{ \log p(\mathbf{x}|\mathbf{z}, c; \boldsymbol{\theta}) \tag{11}$$
$$+ \log p(\mathbf{z}|c) + \log \pi_c \Big\} d\mathbf{z} + \mathcal{H}(q),$$

where $\mathcal{H}(\cdot)$ is the entropy of the posterior distribution $q$ and $\boldsymbol{\theta}$ represent the first layer MFA parameters. The DMFA formulation seeks to find a better prior $\log p(\mathbf{z}|c)$ (using Eq. 10), while holding the first layer parameters fixed. Initially, when $q(\mathbf{z}, c|\mathbf{x}; \boldsymbol{\theta}) \equiv p(\mathbf{z}, c|\mathbf{x}; \boldsymbol{\theta})$, the bound is tight. Therefore, any increase in the bound will lead to an increase in the true likelihood of the model. Maximizing the bound of Eq. 11 with respect to $\boldsymbol{\theta}^{(2)}$ is equivalent to maximizing:

$$\sum_{c=1}^{C} \int_{\mathbf{z}} q(\mathbf{z}, c|\mathbf{x}; \boldsymbol{\theta}) \log p(\mathbf{z}|c; \boldsymbol{\theta}^{(2)}) \tag{12}$$

averaged over the training data vectors. This is equivalent to fitting component-specific second-layer MFAs with vectors drawn from $q(\mathbf{z}, c|\mathbf{x}; \boldsymbol{\theta})$ as data. The same scheme can be extended to training third-layer MFAs. With proper initialization, we are guaranteed to improve the lower bound on the log-likelihood, but the log-likelihood itself can fall (Hinton et al., 2006).

Fig. 1 (middle panel) shows a schematic representation of our model. Using $\pi_{k_c}^{(2)}$ to denote the second layer mixing proportion of component $k_c$, we have:

$$\forall c : \sum_{k_c=1}^{K_c} \pi_{k_c}^{(2)} = 1 \tag{13}$$

A DMFA replaces the old MFA prior $p_{MFA}(\mathbf{z}, c) = p(c)p(\mathbf{z}|c)$ with a better prior:

$$p_{DMFA}(\mathbf{z}, c) = p(c)p(k_c|c)p(\mathbf{z}|k_c) \tag{14}$$

Therefore, when sampling from a DMFA, we first sample $c$ using $\pi_c$, followed by sampling the second layer component $k_c$ using $\pi_{k_c}^{(2)}$. Finally, we can sample $\mathbf{z}$ using the Gaussian of component $k_c$, as in Eq. 4.

A simpler, but completely equivalent DMFA formulation is to enumerate over all possible second layer components $k_c$. We use a new component indicator variable $s = 1, \ldots, S$ to denote a specific second layer component, where $S = \sum_{c=1}^{C} K_c$. The mixing proportions are defined as $\pi_s^{(2)} = p(c(s))p(k_c(s)|c(s))$, where $c(s)$ and $k_c(s)$ denotes the first and second layer components $c$ and $k_c$ to which $s$ corresponds. For example $c(2) = 1$ and $c(5) = 2$. We note that the size of $S$ is exponential in the number of DMFA layers. The generative process of this formulation is very intuitive and we shall use it throughout the remaining sections.

Fig. 1 (right panel) shows the graphical model for a 2 layer DMFA. Specifically,

$$p(s) = \pi_s^{(2)} \tag{15}$$

$$p(\mathbf{z}^{(2)}|s) = \mathcal{N}(\mathbf{z}^{(2)}; 0, \mathbf{I}) \tag{16}$$

$$p(\mathbf{z}^{(1)}|\mathbf{z}^{(2)}, s) = \mathcal{N}(\mathbf{z}^{(1)}; \mathbf{W}_s^{(2)}\mathbf{z}^{(2)} + \boldsymbol{\mu}_s^{(2)}, \Psi_s^{(2)}) \tag{17}$$

$$c \leftarrow c(s), \quad \text{(deterministic)} \tag{18}$$

$$p(\mathbf{x}|\mathbf{z}^{(1)}, c) = \mathcal{N}(\mathbf{x}; \mathbf{W}_c^{(1)}\mathbf{z}^{(1)} + \boldsymbol{\mu}_c^{(1)}, \Psi_c^{(1)}) \tag{19}$$

Eq. 18 is fully deterministic as every $s$ belongs to one and only one $c$. $\mathbf{z}^{(1)} \in \mathbb{R}^{d^{(1)}}$, $\mathbf{z}^{(2)} \in \mathbb{R}^{d^{(2)}}$, $\mathbf{W}_c^{(1)} \in \mathbb{R}^{D \times d^{(1)}}$, $\mathbf{W}_s^{(2)} \in \mathbb{R}^{d^{(1)} \times d^{(2)}}$, $\boldsymbol{\mu}_c^{(1)} \in \mathbb{R}^{d^{(1)}}$, and $\boldsymbol{\mu}_s^{(2)} \in \mathbb{R}^{d^{(2)}}$. Finally, $\Psi_c^{(1)}$ and $\Psi_s^{(2)}$ are $d^{(1)} \times d^{(1)}$ and $d^{(2)} \times d^{(2)}$ diagonal matrices of the first and second layers respectively.

DMFA has an equivalent *shallow* form, which is obtained by integrating out the latent factors. If we integrate out the first layer factors $\mathbf{z}^{(1)}$, we obtain:

$$p(\mathbf{x}|\mathbf{z}^{(2)}, s) = \mathcal{N}(\mathbf{x}; \mathbf{W}_c^{(1)}(\mathbf{W}_s^{(2)}\mathbf{z}^{(2)} + \boldsymbol{\mu}_s^{(2)}) + \boldsymbol{\mu}_c^{(1)},$$
$$\Psi_s^{(1)} + \mathbf{W}_c^{(1)}\Psi_s^{(2)}\mathbf{W}_c^{(1)^\top}) \tag{20}$$

By further integrating out $\mathbf{z}^{(2)}$:

$$p(\mathbf{x}|s) = \mathcal{N}(\mathbf{x}; \mathbf{W}_c^{(1)}\boldsymbol{\mu}_s^{(2)} + \boldsymbol{\mu}_c^{(1)},$$
$$\Psi_c + \mathbf{W}_c^{(1)}(\Psi_s^{(2)} + \mathbf{W}_d^{(2)}\mathbf{W}_d^{(2)^\top})\mathbf{W}_c^{(1)^\top}) \tag{21}$$

From Eq. 20, we can see that a DMFA can be reduced to a standard MFA where $\mathbf{z}^{(2)}$ are the factors and $s$ indicates the mixture component. This "collapsed" MFA is regularized due to its parameter sharing. In particular, the means of the components $s$ with the same first layer component $c$ all must lie on a hyperplane spanned by $\mathbf{W}_c^{(1)}$. The covariance of these components all share the same outer product factorization $(\mathbf{W}_c^{(1)}\mathbf{W}_c^{(1)^\top})$ but with different "core matrices" $(\Psi_s^{(2)} + \mathbf{W}_s^{(2)}\mathbf{W}_s^{(2)^\top})$.

Assuming that the number of the second layer components are equal, i.e. $\forall c : K_c = K$, a standard shallow MFA with $S = C \times K$ mixture components and $d^{(1)}$ factors per component would require $O(DKd^{(1)}C)$ parameters. A DMFA with two layers, on the other hand, would require $O(Dd^{(1)}C + d^{(1)}d^{(2)}CK) = O((D + d^{(2)}K)d^{(1)}C)$ parameters. Note that a DMFA requires a much smaller number of effective parameters than an equivalent shallow MFA, since $d^{(2)} << D$. As we shall see in Sec. 4.1, this sharing of parameters is critical for preventing overfitting.

### 3.1. Inference

Exact inference in a collapsed DMFA model is of order $O(CK)$ since the data likelihood must be computed for each mixture component. We can incur a lower cost by using an approximate inference, which is $O(C + K)$. First, we compute the posterior $p(\mathbf{z}^{(1)}, c|\mathbf{x}) = p(\mathbf{z}^{(1)}|\mathbf{x}, c)p(c|\mathbf{x})$ using Eq. 7. This posterior is exact if we had a standard normal prior over $\mathbf{z}^{(1)}$, but it is an approximation of the exact posterior of the DMFA model. The entropy of the posterior $p(c|\mathbf{x})$ is likely to be very low in high dimensional spaces. We therefore make a point estimate by selecting the component $c$ with maximum posterior probability:

$$\hat{c} = \arg\max_c p(c)p(c|\mathbf{x}) \tag{22}$$

$$p(\mathbf{z}^{(1)}|\mathbf{x}) = \sum_c p(\mathbf{z}^{(1)}|\mathbf{x}, c)p(c|\mathbf{x})dc$$

$$\approx p(\mathbf{z}^{(1)}|\mathbf{x}, \hat{c}) \tag{23}$$

For the second layer, we treat $\hat{c}$ and $\mathbf{z}^{(1)}$ as data, and compute the posterior distribution $p(\mathbf{z}^{(2)}, s|\mathbf{z}^{(1)}, \hat{c})$ in a similar fashion.

### 3.2. Learning

A DMFA can be trained efficiently using a greedy layer-wise algorithm. The first layer MFA is trained in a standard way. We then use Eq. 23 to infer the component $\hat{c}$ and the factors associated with that component for each training case $\{\mathbf{x}_n\}$. We then freeze the first layer parameters and treat the sampled first

---

**Algorithm 1** Learning DMFAs

Given data: $X = \{\mathbf{x}_1, \mathbf{x}_2, \ldots, \mathbf{x}_N\}$.
*//Layer 1 training*
Train 1st layer MFA on $X$ with $C$ components and $d$ factors using EM → MFA1.

*//Layer 2 training*
Create dataset $Y_c$ for each of the $C$ components.
$Y_c \leftarrow \emptyset$
**for** $i = 1$ to $N$ **do**
  **for** $c = 1$ to $C$ **do**
    compute $p(c|\mathbf{x}_i)$ and $p(\mathbf{z}^{(1)}|\mathbf{x}_i, c)$, Eqs. 7 & 8.
  **end for**
  Find $\hat{c} = \arg\max_c p(c|\mathbf{x}_i)$.
  Sample $\mathbf{z}_i^{(1)}$ from $\mathcal{N}(\mathbf{z}^{(1)}; \mathbf{m}_{\hat{c}}, \mathbf{V}_{\hat{c}}^{-1})$.
  Add $\mathbf{z}_i^{(1)}$ to dataset $Y_{\hat{c}}$: $Y_{\hat{c}} = Y_{\hat{c}} \cup \{\mathbf{z}_i^{(1)}\}$.
**end for**

$d^{(2)}$ and $K_c$: # of 2nd layer factors and components.
**for** $c = 1$ to $C$ **do**
  Train a separate 2nd layer MFA on $Y_c$ with $d^{(2)}$ factors and $K_c$ components using EM → MFA2{c}.
**end for**

---

layer factor values for every component $\{\{\mathbf{z}_n^{(1)}\}_c\}$ as training data for the second layer MFAs. Algorithm 1 details this layer-wise training algorithm. After greedy learning, "backfitting" by collapsing a DMFA and running additional EM steps is also possible. However, more care is needed to prevent overfitting.

## 4. Experiments

We demonstrate the advantages of learning DMFAs on both low dimensional and high dimensional datasets, including face images, natural image patches, and speech acoustic data.

**Toronto Face Database** (TFD): The Toronto Face Database is a collection of aligned faces from a variety of (mostly) publicly available face image databases (Susskind, 2011). From the original resolution of $108 \times 108$, we downsampled to resolutions of $48 \times 48$ or $24 \times 24$. We then randomly selected 30,000 images for training, 10,000 for validation, and 10,000 for testing.

**CIFAR-10**: The CIFAR-10 dataset (Krizhevsky, 2009) consists of 60,000 $32 \times 32 \times 3$ color images of 10 object classes. There are 50,000 training images and 10,000 test images. Out of 50,000 training images, 10,000 were set aside for validation.

**TIMIT Speech**: TIMIT is a corpus of phonemically and lexically transcribed speech of American English speakers of different sexes and dialects[1]. The corpus contains a 462-speaker training set, a 50-speaker validation set, and a 24-speaker core test set. For our purposes, we extracted data vectors every 10-ms from the continuous speech data. Each frame analyzes a 25-ms Hamming window using a set of filter banks based on the Fast Fourier Transform. Concatenating 11 frames, we obtain 1353 dimensional input vectors. We randomly selected 30,000 vectors for training, 10,000 for validation, and 10,000 for testing.

**Berkeley Natural Images**: The Berkeley segmentation database (Martin et al., 2001) contain 300 images from natural scenes. We randomly extracted 2 million $8 \times 8$ image patches for training, 50,000 patches for validation, and 10,000 for testing.

**UCI**: We used 4 datasets from the UCI repository (Murphy & Aha, 1995). These are low dimensional datasets and have relatively few training examples. These were the only UCI datasets we tried.

For all image datasets, the DC component of each image was removed: $\mathbf{x} \leftarrow \mathbf{x} - mean(\mathbf{x})$. This removes the huge illumination variations across data samples. No other preprocessing steps were used. For the TIMIT and UCI datasets, we normalize input vectors to zero mean and scale the entire input by a single number to make the average standard deviation be one. For evaluating the log probabilities of DMFAs, we always first collapsed it to a shallow MFA in order to obtain the exact data log-likelihood.

### 4.1. Overfitting

We first trained a 20 component MFA on $24 \times 24$ faces until convergence[2], which took 33 iterations. The number of factors was set to half of the input dimensionality, $d^{(1)} = D/2 = 288$. Fig. 2 shows the corresponding training and validation log-likelihoods[3]. We next stacked a second MFA layer with five second layer components ($K_c = 5$) for each of the first layer components and $d^{(2)} = 50$ second layer factors. The DMFA (MFA2) model improved as learning continued for an additional 20 iterations (see red and blue lines in Fig. 2). As a comparison, immediately after we initially formed the two-layer MFA, we collapsed it into its equivalent shallow representation and performed additional training (magenta and black lines in Fig. 2). Observe that the shallow MFA starts overfitting due to its extra capacity (5 times more parameters). MFA2, on the other hand, shows improvements on both the training and validation data. We note that training a shallow MFA with 100 components from random initialization is significantly worse (see Table 1).

---

[1] www.ldc.upen.edu/Catalog/CatalogEntry.jsp?catalogId=LDC93S1
[2] Convergence is achieved when the log-likelihood changed by less than 0.01% from the previous EM iteration.
[3] Similar results were obtained for different numbers of components and factors.

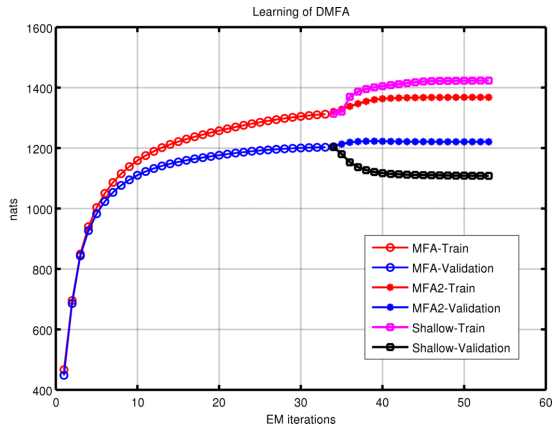

*Figure 2.* DMFA improves over MFA. Overfitting occurs during further training of a shallow MFA with increased capacity. Best viewed in color.

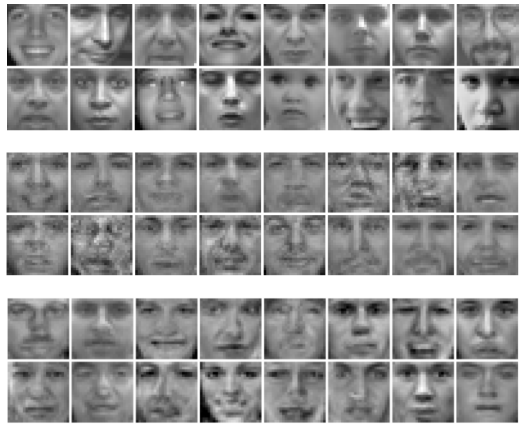

*Figure 3.* **Top**: training images. **Middle**: samples from MFA. **Bottom**: samples from DMFA.

To give a sense of the computation costs, training the first layer MFA took 1600 seconds on a multi-core Xeon machine. The second layer MFA training took an additional 580 seconds.

### 4.2. Qualitative Results

We next demonstrate qualitative improvements of the samples from a DMFA over a standard MFA model. As the baseline, we first trained a MFA model on 30,000 $24 \times 24$ face images from the TFD, with 288 factors and 100 components. We then trained a DMFA with 20 first layer components and 5 second layer components for each of the 20 first layer components. The DMFA has the same number of parameters as the baseline MFA. The two-layer MFA (MFA2) performs better compared to the standard MFA by around 20 nats on the test set. Fig 3 further shows samples from the two models. Qualitatively, the DMFA appears to generate better samples compared to the shallow MFA model.

### 4.3. High Dimensional Data

Next, we explore the benefits of DMFAs on the high dimensional CIFAR and TIMIT datasets. We first trained a MFA model with the number of factors equal to half of the input dimensionality. The number of mixture components was set to 20. For MFA2, 5 components with 50 latent factors were used. For the 3rd layer MFA (MFA3) 3 factors with 30 latent factors were used.

Table 1 shows the average training and test log-likelihood. In addition, we provide results for two types of RBM models that are commonly used when modeling high-dimensional real-valued data, including image patches and speech. The SSU model is a type of RBM with Gaussian visible and stepped sigmoid hidden units (Nair & Hinton, 2010). By using rectified linear activations, SSU can be viewed as a mixture of linear models with the number of components exponential in the number of hidden variables. A simpler Gaussian RBM (GRBM) model uses Gaussian visible and binary hidden units. It can also be viewed as a mixture of diagonal Gaussians with exponential number of components. For both the GRBM and SSU, we used Fast Persistent Contrastive Divergence (Tieleman & Hinton, 2009) for learning and AIS (Salakhutdinov & Murray, 2008) to estimate their log-partition functions. The AIS estimators have standard errors of around 5 nats, which are too small to affect the conclusions we can draw from Table 1.

The number of parameters for the GRBMs and SSU are matched to the MFA model, which means that approximately 6,000 hidden nodes are used. Increasing the number of hidden units did not result in any significant improvements of GRBM and SSU models. Hyperparameters are selected using the validation set. After MFA learning converged, a MFA2 model is initialized. The means of the MFA-2 components were slightly perturbed from zero so as to break symmetry. Shallow1 results were obtained by collapsing these newly initialized MFA2 models and further training using EM with early stopping. Shallow2 results were obtained by starting at random initialization (with multiple restarts) with the equivalent number of parameters as the corresponding Shallow1 models. We note the significant gains by DMFAs for the TIMIT and TFD-48 datasets.

Fig. 4 displays gains of 2 and 3 layer MFA as we vary the number of the first layer mixture components. It is interesting to observe that MFA and DMFA significantly outperformed various RBM models. This result suggests that it may be possible to improve many of the existing deep networks for modeling real-valued data that use GRBMs for the first hidden layer, though better density models do not necessarily learn features that are better for discrimination.

| Dataset | GRBM | SSU | MFA | MFA-2 | MFA-3 | Shallow1 | Shallow2 | Diff-2 | Diff-3 |
|---|---|---|---|---|---|---|---|---|---|
| TFD-24 | 766 | 859 | 1312 | 1368 | 1380 | 1325 | 1506 | 57.1 ± 0.1 | 12± 0.2 |
| | **758** | **841** | **1185** | **1202** | **1207** | **1184** | **1039** | **18.7 ± 0.2** | **4.1 ± 0.08** |
| TFD-24-Rot | 843 | 950 | 1412 | 1469 | 1477 | 1428 | 1505 | 56.9 ± 0.1 | 8.5 ± 0.13 |
| | **822** | **929** | **1283** | **1305** | **1306** | **1284** | **1125** | **21.6 ± 0.2** | **1.4 ± 0.04** |
| TFD-48 | 2426 | 3675 | 6020 | 6141 | 6151 | 6036 | 6461 | 119.2 ± 0.4 | 11.7 ± 0.1 |
| | **2413** | **3557** | **5159** | **5242** | **5250** | **5161** | **4299** | **85.3 ± 0.5** | **5.6 ± 0.1** |
| CIFAR: | 2725 | 2818 | 4486 | 4573 | 4583 | 4565 | 4214 | 86.8 ± 0.4 | 10.6 ± 0.2 |
| | **2365** | **2494** | **3587** | **3621** | **3622** | **3592** | **2873** | **33.2 ± 0.4** | **0.5 ± 0.06** |
| TIMIT: | 1244 | 1316 | 2662 | 2802 | 2804 | 2707 | 3219 | 133.5 ± 0.2 | 1.4 ± 0.05 |
| | **1175** | **1268** | **2298** | **2450** | **2451** | **2305** | **1169** | **147.2 ± 0.5** | **0.4 ± 0.07** |

*Table 1.* Model performance on various high dimensional datasets (nats). TFD-24-Rot is generated by randomly rotating $24 \times 24$ face images in the range of $\pm 45$ deg. Diff-2 and Diff-3 are the gains from going from 1 to 2 layers and from 2 to 3 layers, respectively. For all datasets, Diff-2 and Diff-3 are statistically significant at $p = 0.01$.

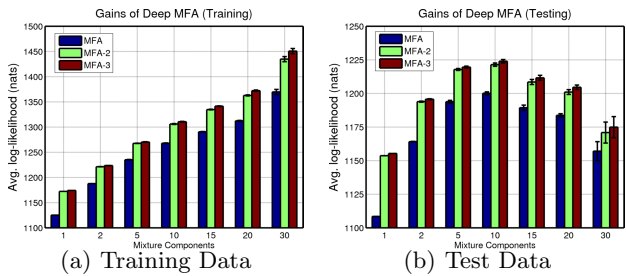

(a) Training Data     (b) Test Data

*Figure 4.* Improvements of DMFA over standard MFA on $24 \times 24$ face images vs. the number of first layer components. Gains are observed across different numbers of first layer components. Surprisingly, while the dataset contains thousands of different people, more than 10 mixture components results in overfitting. Best viewed in color.

### 4.4. Low Dimensional Data

DMFAs can also be used with low dimensional data. Following (Silva et al., 2011), we used 4 continuous datasets from the UCI repository. We removed the discrete variables from all datasets. For the Parkinsons dataset, one variable from any pair whose Pearson correlation coefficient is greater than 0.98 was also removed (for details see (Silva et al., 2011)). Table 2 reports the averaged test results using 10-fold cross validation. Compared to the recently introduced Copula Networks, MFAs give much better test predictive performance. Adding a second layer produced significant gains in model performance. The improvements from adding a second layer on all datasets were statistically significant using the paired t-test at $p = 0.01$.

| PIX | PCA | ICA | GMM | MFA | MFA-2 |
|---|---|---|---|---|---|
| 78.3 | 114.2 | 115.9 | 167.2* | 166.5 | 169.3 |

*Table 3.* Average test log-likelihood (in nats) of various models learned on $50,000$ $8 \times 8$ test patches. **PIX**: independent pixels. **PCA**: Principle Component Analysis. **ICA**: Independent Component Analysis. **GMM**: Mixture of Gaussians with 200 components. **MFA** Mixture of Factor Analysers with 200 components. **MFA-2** Two layer DMFA. MFA and MFA-2 results are from our experiments, other numbers are taken from Zoran & Weiss (2011). GMM's 167.2* is different from the previously reported 164.5 due to the random extraction of test patches. 167.2 was obtained by evaluating the downloaded model of Zoran & Weiss (2011) on our own test patches.

### 4.5. Natural Images

One important application of generative models is in the task of image restoration which can be formulated as a MAP estimation problem. As confirmed by Zoran & Weiss (2011), a better prior almost certainly leads to a better signal to noise ratio of the restored image. In addition, Zoran & Weiss (2011) have shown that combining a mixture of Gaussians model trained on $8 \times 8$ patches of natural images with a patch-based denoising algorithm, allowed them to achieve state-of-the-art results. Following their work, we trained a two-layer MFA on $8 \times 8$ patches from the Berkeley database. Two million training and 50,000 test patches were extracted from the 200 training and 100 test images, respectively. Table 3 shows results. Note that the DMFA improves upon the current state-of-the-art GMMs model of Zoran & Weiss (2011) by about 2 nats, while substantially outperforming other commonly used models including PCA and ICA. Finally, we trained a shallow equivalent to MFA-2 (5 times more parameters than MFA) from random initialization and achieved only 164.9 nats, thereby demonstrating that DMFAs are necessary in order to achieve the extra gain.

| Dataset | dim. | size | Gaussian | Cop. MCDN | MFA | MFA-2 | DMFA gain |
|---|---|---|---|---|---|---|---|
| Parkinsons | 15 | 5875 | -11.65 | -3.48 | -0.63 | -0.33 | $0.296 \pm 0.024$ |
| Ionosphere | 32 | 351 | -41.10 | -27.45 | -20.10 | -18.53 | $1.565 \pm 0.252$ |
| Wine(red) | 11 | 1599 | -13.72 | -11.25 | -10.22 | -10.07 | $0.143 \pm 0.015$ |
| Wine(white) | 11 | 4898 | -13.76 | -12.11 | -11.02 | -10.89 | $0.121 \pm 0.036$ |

*Table 2.* Test set predictive log-likelihood on 4 UCI datasets (nats). Reported results are from 10-fold cross validation on each dataset. MFA results are from our experiments. Other results are from (Silva et al., 2011).

### 4.6. Allocating more components to more popular factor analysers

Until now, we have given every higher level MFA the same number of components to model the aggregated posterior of its lower level factor analyser ($\forall c : K_c = K$). While simple to implement, this is not optimal. An alternative is to use more second layer components for the first layer components with bigger mixing proportions. We tested this hypothesis by first training a MFA model on $48 \times 48$ TFD faces, which achieved an average test log-likelihood of 5159 nats. For the two-layer MFA, instead of assigning 5 components to each of the first layer components, we let $K_c \propto \pi_c$, with $min(K_c) = 2$ and $\sum_c^C K_c = 5 \times C$. With all other learning hyper-parameters held constant, the resulting DMFA achieved 5246 nats on the test set. Compared to 5242 nats of our previous model (c.f. Table 1), the new method accounted for a gain of 4 nats. As another alternative, a measure of Gaussianity of the aggregated posterior could be used to determine $K_c$.

## 5. Discussions

As density models, MFAs significantly outperform undirected RBM models for real-valued data and by using second layer MFAs to model the aggregated posterior of each first layer factor analyser, we can achieve substantial gains in performance. Higher input dimensionality leads to bigger gains from learning DMFAs. However, adding a third MFA layer appears to be of little value. Another possible extension of our work is to train a mixture of linear dynamical systems and then to train a higher-level mixture of linear dynamical systems to model the aggregated posterior of each component of the first level mixture.

## Acknowledgements

We thank Iain Murray for discussions and Jakob Verbeek for sharing his MFA code. This research was supported by NSERC & CIFAR.